\documentclass{article}
\usepackage{geometry}

\usepackage{graphics} 
\usepackage{epsfig} 
\usepackage{mathptmx} 
\usepackage{times} 
\usepackage{amsmath} 
\usepackage{amssymb}  
\usepackage{graphicx}
\usepackage{booktabs}
\usepackage{hyperref}
\usepackage{url}
\usepackage{bbm}
\usepackage{cleveref}
\usepackage{subfig}
\usepackage{overpic} 
\usepackage{algpseudocode}

\title{\LARGE \bf
Learning Complicated Manipulation Skills via Deterministic Policy with Limited Demonstrations
}

\author{Haofeng Liu$^{1,2}$, Jiayi Tan$^{1,2}$, Yiwen Chen$^{1,2}$ and  Marcelo H Ang Jr$^{1,2}$
\thanks{$^{1}$Department of Mechanical Engineering, National University of Singapore}%
\thanks{$^{2}$Advanced Robotics Centre, College of Design and Engineering, National University of Singapore}
}

\date{}

\begin{document}
\maketitle

\begin{abstract}
\normalsize 
Combined with demonstrations, deep reinforcement learning can efficiently develop policies for manipulators. However, it takes time to collect sufficient high-quality demonstrations in practice. And human demonstrations may be unsuitable for robots. The non-Markovian process and over-reliance on demonstrations are further challenges. For example, we found that RL agents are sensitive to demonstration quality in manipulation tasks and struggle to adapt to demonstrations directly from humans. Thus it is challenging to leverage low-quality and insufficient demonstrations to assist reinforcement learning in training better policies, and sometimes, limited demonstrations even lead to worse performance.

We propose a new algorithm named TD3fG (TD3 learning from a generator) to solve these problems. It forms a smooth transition from learning from experts to learning from experience. This innovation can help agents extract prior knowledge while reducing the detrimental effects of the demonstrations. Our algorithm performs well in Adroit manipulator and MuJoCo tasks with limited demonstrations.
\end{abstract}

\section{INTRODUCTION}

 While solving individual tasks via manipulators in a controlled setting has led to success in industrial automation, this is less feasible in unstructured settings like the home. Deep Reinforcement Learning (DRL) help manipulator adapt to various environments by interacting with the environment. It has been proven successful in many domains, exceeding human performance in video games \cite{mnih2015human}, robot manipulators\cite{2018Learning}, and various open-source simulations\cite{2015Continuous}. However, DRL demands frequent interactions, which proves challenging in manipulator scenarios. Robots may struggle with suboptimal policies and spend time exploring until they accumulate valuable experience, resulting in low sampling efficiency and slow convergence\cite{wang2021policy}.

One approach to addressing these limitations is leveraging human expert demonstrations\cite{goecks2019integrating}. In DRL with demonstrations, the agent can learn from both expert behavior and its experience enhancing manipulators' performance in terms of generalization and adaption to unstructured environments\cite{brys2015reinforcement}\cite{codevilla2019exploring}. 

Nevertheless, reinforcement learning with demonstrations faces hurdles, such as the demonstration's lack of Markov properties and unrepresentable behaviors. Particularly with limited poor demonstrations, the performance tends to decline\cite{ross2011reduction}. Furthermore, obtaining sufficient samples in practice is often infeasible or prohibitively expensive\cite{wang2021policy}. Over-reliance on demonstrations may yield unfavorable results\cite{wu2019behavior}. Demonstrations from hand-crafted deterministic policies or rise in human behavior may cause divergence both empirically\cite{fujimoto2019off} and theoretically \cite{dulac2019challenges} \cite{fu2020d4rl}. Previous methods like SACfD and DDPGfD\cite{vecerik2017leveraging} expose these issues in our manipulator experiments; thus, we anticipate agents to boost their performance from sub-optimal, even failed data.

We propose TD3 learning from the generator (TD3fG) to solve the above problems. Our methods utilize demonstrations to modify the exploration noise and loss function instead of applying pre-train and experience replies. The results of the comparison experiment exhibit a significant improvement in training efficiency and final performance. To summarize, the main contribution of this work:

$\bullet$ Propose the TD3fG algorithm in section IV. It modifies demonstrations in action noise and loss functions and deploys a smooth transition to avoid overdependence.

$\bullet$ In section VI, we sample 100 demonstrations and show comparison experiments. The results indicate our approach is better adapted to limited demonstrations in complex manipulator tasks.

$\bullet$ In section VII, we perform ablation experiments to investigate the contributions of the individual components in our method.

\section{RELATED WORK}

\subsection{Reinforcement Learning}

Researchers have widely applied DRL in auto-driving and robotics and highly investigated it because of its high-level autonomy\cite{silver2016mastering}\cite{levine2016end}. A renewed interest in RL cascaded from success in bipedal robots and manipulator control \cite{gu2017deep}. 

In RL, we assume finishing a task is a Markov Decision Process (MDP) defined by a tuple 
$ (S, A, p, r, \gamma)$ where $S$ and $A$ are sets of states and actions. $p(s_0|s, a)$ is the state transfer probability. $r:S \times A \to R$ is a reward function and $\gamma \in [0, 1)$ is the discount factor. The Markovian property of MDP indicates the conditional probability distribution of the future state only depends on the present state\cite{boyen2013tractable}.

For the Actor-Critic framework\cite{andrychowicz2021matters}, the actor selects actions with the max Q-value to maximize the total reward. The Q value from the critic represents the expectation of future accumulated rewards. It is expressed as $Q_t=\sum_{i=t}^T\gamma^{i-t}r_i$.

\begin{equation}
\begin{aligned}
    Q_\pi(s_t,a_t) &= {E}_{r_t,s_t\sim E, a_i\sim \pi}[Q^\pi|s_t,a_t]\\
    &= {E}_{r_t,s_{t+1}\sim E, a_i\sim \pi}[r_t \\
    &+\gamma{E}_{a_t\sim\pi}[Q^\pi(s_{t+1}, a_{t+1}]]
\end{aligned}
\end{equation}

\subsection{Imitation Learning}

Behavior cloning (BC) is one of the simplest forms of imitation learning. It updates the policy network to minimize the mean square error (Deterministic) or cross-entropy (Stochastic) between the output and target actions. For deterministic policy, it is formalized as:
\begin{equation}
\label{equ:train generator}
    L(\phi_t)=\frac{1}{N}\sum_i(a_{demo}-\pi_{\phi_t}(s_{demo})^2
\end{equation}

BC has achieved great success in driving, locomotion, and navigation but still struggles with limited demonstrations \cite{8463162}. The bias from experts' track and interference of poor trajectory seriously affects the performance, as observed in several prior works \cite{fujimoto2019off}\cite{kumar2019stabilizing}\cite{levine2020offline}\cite{sutton2018reinforcement}. This weakness becomes obvious in manipulator tasks with high-dimension action and state space.

Data Set Aggregation (DAgger) interleaves between expert and learned policy to address the problem of accumulating error \cite{ross2011reduction}. Furthermore, Safe DAgger introduces a safety policy to predict the error from the primary policy\cite{zhang2016query}. However, DAgger requires access to experts to fix the accumulated errors, which makes it less feasible \cite{kiran2021deep}.

\subsection{Reinforcement Learning with Demonstrations}

Previous works have integrated expert demonstrations to accelerate the training, like DQfD\cite{codevilla2019exploring}, SACfD\cite{haarnoja2018soft}, and DDPGfD\cite{2017arXiv170708817V}. DQfD imports a margin loss which guarantees the expert actions have higher Q-values than other actions. In DDPGfD, the authors import demonstrations into the replay buffer and introduce new methods to incorporate demonstrations.

DQfD, DDPGfD, and SACfD maintain a replay buffer to store demonstrations and experience and draw extra $N_D$ examples from $R_D$. In further work, researchers introduce a behavior cloning loss only on the demonstration samples\cite{8463162}. It can omit the pre-training and learn from the demonstration while interacting with the environment. The limitation is sample inefficiency and the need for a large amount of experience, which is impractical for manipulator tasks.

However, the above methods are sensitive to the quantity and quality of the demonstrations. As shown in Section VI, when only providing a small batch of human samples, they cannot get good results as in \cite{haarnoja2018soft}\cite{2017arXiv170708817V}\cite{codevilla2019exploring}

\subsection{Twin Delayed Deep Deterministic policy gradient}
Twin Delayed Deep Deterministic policy gradient algorithm (TD3) is a deterministic off-policy algorithm that excels in continuous action control\cite{fujimoto2018addressing}. We adopt it as the base algorithm for the RL part and combine it with another approach to learn from demonstrations and control the weight for the demonstration parts.

It alternately interacts with the environment and updates its networks\cite{lillicrap2015continuous}. For every step, the agent executes a selected action from $a_t=\pi_{\phi'}(s_t)+\mathcal{N}_t$ according to policy $\pi$ and exploration noise $\mathcal{N}$. During the training, the agent samples a random mini-batch $N$ from replay buffer $R$, then updates the critic according to the Bellman equation and updates the actor to maximize the Q value:
\begin{equation}
    \begin{aligned}
    &y_i=r_i+\gamma {min}_{i=1,2}Q_{\theta'_i}(s_{i+1}, \pi_{\phi'}(s_{i+1}))\\
    &\theta_i \leftarrow \mathrm{argmin}_{\theta_i} N^{-1} \sum (y - Q_{\theta_i}(s,a))^2\\ 
    &\nabla_{\phi} J(\phi) = N^{-1} \sum \nabla_{a} Q_{\theta_1}(s, a) |_{a=\pi_{\phi}(s)} \nabla_{\phi} \pi_\phi(s)
    \end{aligned}
\end{equation}

\section{PROBLEM FORMULATION}
\begin{figure}[t]
    \centering
    \includegraphics[scale=0.5]{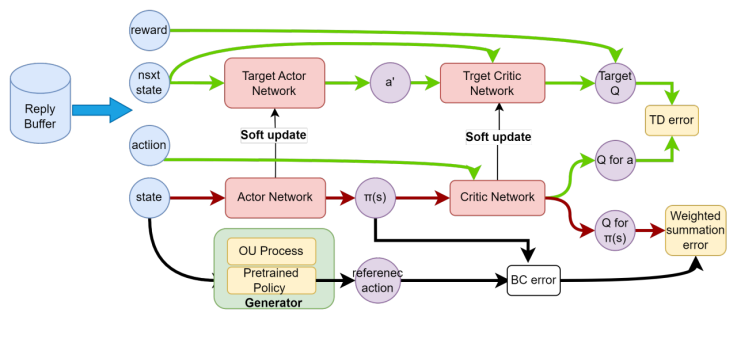}
    \caption{the flowchart of TD3fG, where the action from generator performs as ground truth in the supervised learning part to calculate the behavior cloning loss $L^{G}$. The green trajectory calculates the TD error for the Critic network, and the red and black trajectory gets the weighted error for the actor network.}. 
    \label{fig:BackTD3fG}
\end{figure}

\begin{figure}[t]
    \centering
    \includegraphics[scale=0.5]{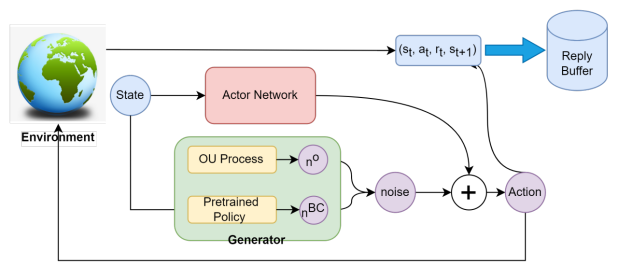}
    \caption{The forward progress of TD3fG, where the generator outputs a combination of reference action from an estimated human policy and a random noise from the OU process. The action noise contains prior knowledge based on demonstrations and could guide the agent's exploration }
    \label{fig:Forward TD3}
\end{figure}
Our problem concerns policy learning for manipulation with limited demonstrations, which has a low average score and is mixed with failed trajectories or human samples that do not fit manipulations well. We focus on tasks that represent a significant fraction of tasks required in our daily life, e.g., tool use and manipulating environmental props\cite{2018Learning}. Developing a robust policy with complicated skills needs long-term training. The introduction of demonstrations can facilitate training, but in practical applications, it could be expensive or impractical to collect large-scale, high-quality samples for robots. Our environment comprises dexterous manipulation and interactive objects. We leverage the demonstrations in two ways, committed exploration and imitation, and look into the possible adverse effects of demonstration. By gradually transforming attention from demonstrations to experience, our methods successfully keep efficient training and achieve the best performance in manipulation tasks, as shown in \cref{Experiment}.

In this work, we tackle the problem under the setting of reinforcement learning with demonstrations, where the demonstrations are from humans (Adriot) or early-stopped RL agents (MuJoCo) and only 100 trajectories for each task, including suboptimal and failed samples. After achieving noticeable improvement in manipulator tasks, we extend our approach to other robotics tasks (gym MuJoCo)\cite{1606.01540} and do ablation studies to analyze the contribution of different components of the method.

\section{METHOD}
\label{Method}

There are two considerations for importing demonstrations, accelerating training and avoiding poor trajectories. We utilize demonstrations in action noise and imitation loss. At early steps imitating the generator. 's actions can help the agent shrink the exploration scope. 

\subsection{Pre-trained reference action generator}

First, the expert demonstrations are used to train a policy neural network $G_\phi$ through supervised learning. We only prepare 100 trajectories for every task, including sub-optimal and even failed trajectories. \cref{table:demonstration details} shows the max, min, and mean scores of demonstrations. 
\begin{equation}
    L(\phi) =\frac{1}{N}\sum_i(a_{demo}-G_{\phi}(s_{demo})^2
\end{equation}
where $a_{demo}$ and $s_{demo}$ are actions and state from the demonstrations.

\subsection{Generate exploration noise}
The modified exploration noise consists of Ornstein Uhlenbeck noise (from the OU process)\cite{gillespie1996exact} and references action noise (from the generator).

The OU noise can improve the exploration efficiency of control tasks in inertial systems and help explore environments with momentum. The noise is presented as $\mathcal{N}$ in $a \leftarrow \pi_\phi(s) + \mathcal{N}$. The differential of OU noise is:
\begin{equation}
\begin{aligned}
    &dn^o(t) = -\zeta(x_t-\mu)dt+d\sigma W_t\\
    &W_t-W_s\sim\mathcal{N}(0, \sigma^2(t-s))\\
    &n^o(t) = \mu + (x_0 - \mu)e^{-\zeta t}
\end{aligned}
\end{equation}
where $\mu$ is mean value, $W_t$ is wiener process. When the state deviates, the noise will pull it close to the mean. The increment in each time interval is the Gaussian distribution. $\sigma^2$ represents the variance of a Wiener process, and it determines the magnification of the disturbance.

The second part from the generator is an unstable but meaningful reference action, providing a more explicit exploration direction than random exploration actions. For example, over half of the trajectories in the Hammer fail to drive the nail with the tool, but they provide actions like moving the manipulator close or even picking up the tool. The reference action is added as a bias to the noise to guide committed exploration.
\begin{equation}
\begin{aligned}
    & a_t \sim \mathcal{N}(a_t, \sigma^2)\\
    & \hat{a_t}\sim \mathcal{N}(a_t + \alpha(t)G(s_t),  \sigma^2)
\end{aligned}
\end{equation}
\indent The weights are decreased along with training to make agents gradually explore conservatively and exploit experience more. We set $T_1 = T_2 = 0.5T_{max}$ to simplify tuning, where $T_max$ is the max training steps.
\begin{equation}
\label{equ:action noise}
\begin{aligned}
    &n(t, s_t) = \alpha(t)n^o(t) + \beta(t)G(s_t)\\
    &\alpha(t) = max(1 - \frac{t}{T_1},0)\\
    &\beta(t) = max(1 -\frac{t}{T_2},0)
\end{aligned}
\end{equation}

\subsection{Generate BC loss}

Third, we import a BC loss between output actions and reference actions. At early steps imitating the actions can help the agent shrink the exploration scope.
\begin{equation}
\begin{aligned}
    & a_t^{ref} = G(s_t)\\
    & L^{G} = (a_t^{ref} - \pi_{\phi_t}(s_t))^2
\end{aligned}
\end{equation}
 $L^{BC}$ has a linear decreasing weight $\epsilon(t)$ and the policy gradient has an increasing weight $\delta(t)$. It allows the agent to switch from imitation to exploitation gradually.
\begin{equation}
\label{equ:loss}
\begin{aligned}
    &L = \epsilon(t)L^{G}(a_t) - \delta(t)Q_{\theta_t}(s_t, a_t)|_{a_t=\pi_{\phi_{t}}(s_t)}\\
    &\epsilon(t) = max(1-\frac{t}{T_{max}},0)\\
    &\delta(t) = min(1 -\epsilon(t),1)
\end{aligned}
\end{equation}
From the aspect of restriction, we expect the agent to select actions with maximum Q value meanwhile stay within a safe distance with the human policy to avoid blind exploration.
\begin{equation}
\begin{split}
        &\pi_{k+1} = \mathop{\mathrm{argmax}}\limits_{\pi \in \Pi} \mathbb{E}_{a\sim\pi(\cdot|x)}[Q(s_t, a_t)],\\
        &\mathop{s.t.} {||\pi_k(s_t)-\pi^{G}(s_t)||}^2 < f(t)
\end{split}
\end{equation}
The safe distance $f(t) = \frac{T_{max}}{T_{max}-t}$ varies from 1 to $+\infty$, where $T_{max}$ is the max training steps, and $t$ is the current number of steps. At the start, $f(t)=0$, there is a strict restriction to guide exploration, which keeps agent policy close to human policy. In the end, $f(t)=+\infty$ indicates the restriction is completely loosened, and the agent entirely concentrates on maximizing the Q value. 
\begin{equation}
\label{equ:Restriction}
\begin{split}
        &\pi_{k+1} = \mathop{\mathrm{argmin}}\limits_{\pi \in \Pi} \mathbb{E}_{a\sim\pi(\cdot|x)}[-Q(s_t, a_t)],\\
        &\mathop{s.t.} {||\pi_k(s_t)-\pi^{G}(s_t)||}^2 \frac{(T_{max} - t)}{T_{max}} < 1
\end{split}
\end{equation}
Through Lagrange multipliers, we can derive the Lagrange:
\begin{equation}
\begin{split}
        &L(\phi_t,\lambda) = \lambda \epsilon(t)L^{G}(a_t)-Q_{theta_t}(s_t, a_t)|_{a_t=\pi_{\phi_t}(s_t)}-\lambda
\end{split}
\end{equation}
The Lagrangian form is very close to the weighted lost function in \cref{equ:loss} except an extra $\lambda$. Finding Lagrangian's local minimal is similar to calculating the gradient of the loss function.

\section{EXPERIMENT SETUP}
We execute the experiments in Adriot manipulation tasks Hammer and Door, which are considered two of the most difficult tasks in this domain. These are typical manipulation tasks with tool use and manipulating environmental props, representing the necessary skills required for our daily robot assistance. Then we expand our tasks to gym MuJoCo tasks, which stands for the robot motion control in a simple environment.
\subsection{Environments}
\begin{figure}[t]
\centering
    \subfloat[Hammer]{\includegraphics[height=0.18\textwidth]{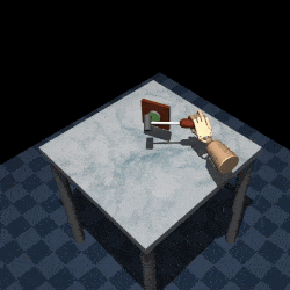}} \ 
    \subfloat[Door]{\includegraphics[height=0.18\textwidth]{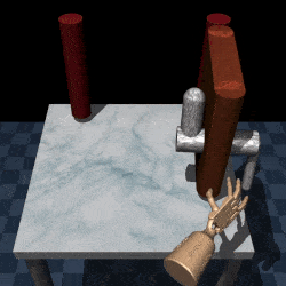}}
    \label{fig:my_label}
\end{figure}

We evaluate our methods in Adroit tasks (Hammer and Door), then extend to Mujoco simulations (Ant, Walker2d, and HalfCheetah\cite{1606.01540}). To meet our assumptions that the demonstration is limited, we only sample 100 samples with low average scores. The samples for Adriot are from VR equipment that records human behavior. For MuJoCo tasks, the dataset is generated by a partially trained Soft Actor-Critic policy.
\begin{table}[h]
\label{table:demonstration details}
\caption{Demonstration Details}
\label{table_time}
\scalebox{0.8}{
\begin{tabular}{cccccc}  
\toprule   
     & Ant & HalfCheetah &Walker2d &Hammer &Door\\  
\midrule 
    Average   &4635.60 &1945.72 &2872.05 &1096.82 &397.83\\
    Max Score       &5487.11 &4354.80 &3914.03 &9570.42 &1136.83 \\
    Min Score       &-685.92 &-638.49 &-5.69 &-260.24 &-51.03\\
    Std             &1356.02 &1586.57 &1066.74 &2286.31 &338.91\\
  \bottomrule  
\end{tabular}
}
\end{table}

\subsection{Network architecture}

The network structure is illustrated in Fig 1. The reference action from the generator affects backpropagation, as shown in Fig 2. In Fig 1, the green arrows calculate the TD error for the Critic network, and the red and black trajectories get the weighted error for the actor network. The red course is the process of exploitation and maximizing Q value. The black course is for imitation based on the distance between the agent's action and the estimated human action. Fig 2 indicates how the agent interacts with the environment. At every step, it executes an action based on the actor's output and the noise. A random OU process noise and a reference action from a pre-trained policy network constitute the noise.

\section{EXPERIMENTAL RESULT}
\label{Experiment}

\subsection{Tasks}
Hammer: The manipulator has to use a hammer to drive the nail into the board. The nail is randomly positioned and with significant dry friction.

Door: The manipulator needs to undo the latch and open the door at a random location. The agent should leverage environmental interaction to develop an understanding of the latch.

Ant, Walker2d, and HalfCheetah: Control a two-dimensional bipedal/quadruped robot walk forward.

\subsection{Baselines}

BC+Finetunes: Directly put the generator in simulation and fine-tune it.

DDPGfD SACfD: They pre-train the model and store the demonstrations in the replay buffer\cite{2017arXiv170708817V}.

TD3: Original TD3 without demonstration information.

\subsection{Results}

Total reward indicates the level of task completion. Fig 3 and Fig 4 compare our work with the TD3, BC + fine-tuning, SACfD, and DDPGfD. 

\begin{figure}[h]
    \centering
    \subfloat[Hammer]{\includegraphics[height=0.18\textwidth]{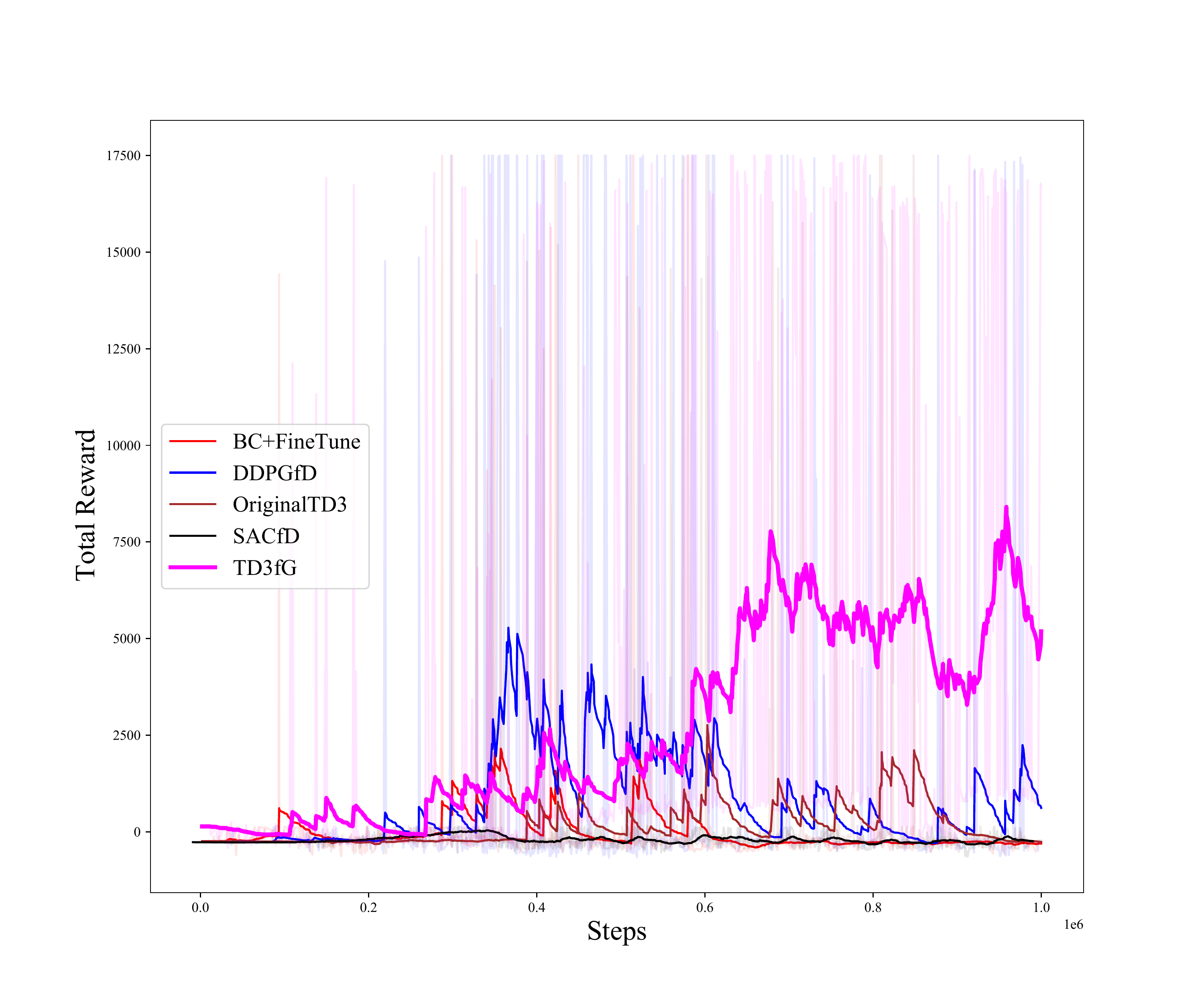}}
    \subfloat[Door]{\includegraphics[height=0.18\textwidth]{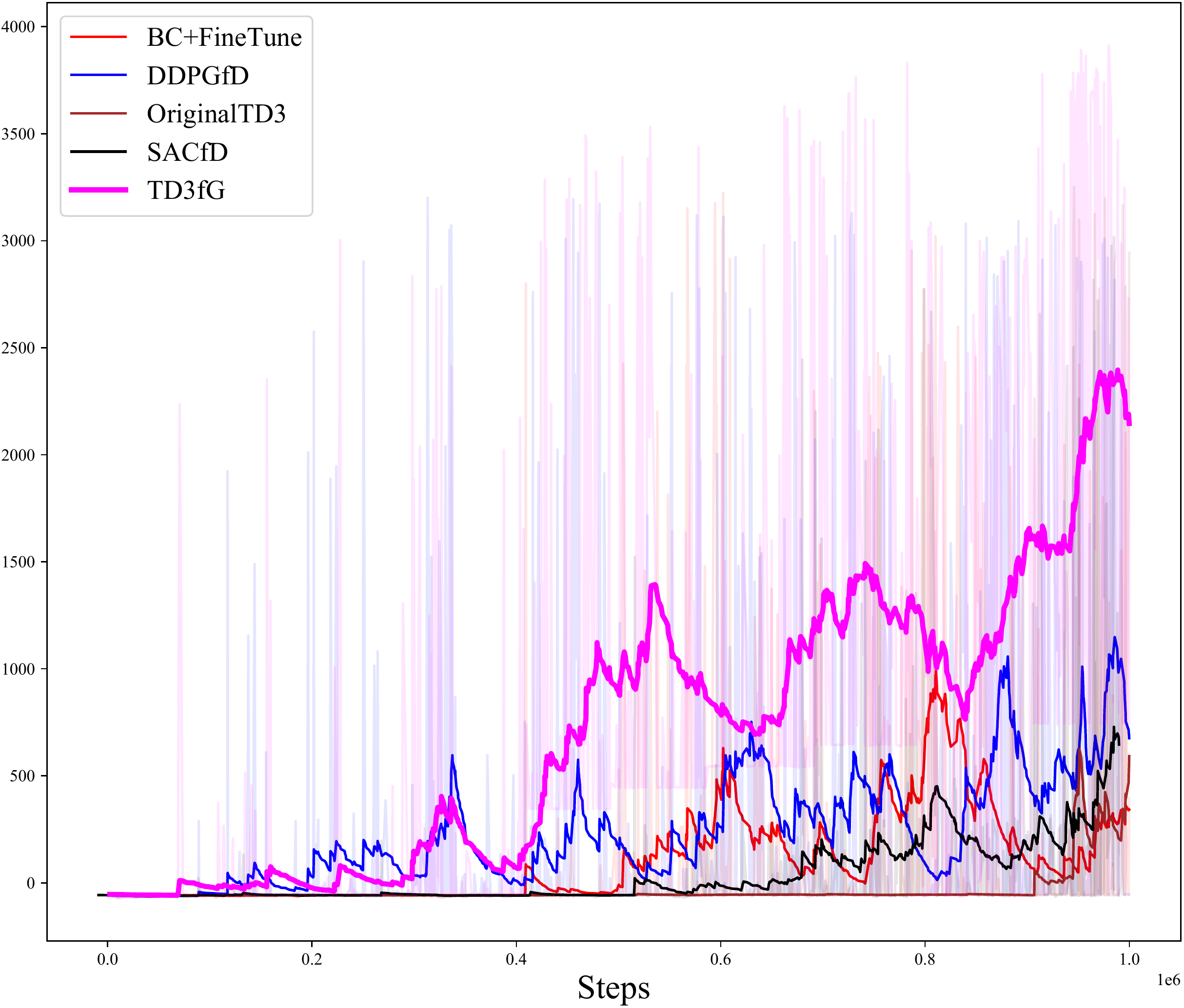}}  
    \caption{Experiment results for Adriot Manipulator tasks, where each curve presents the average total reward of 6 random seeds}
    \label{fig:Compararison}
\end{figure}
Fig 3 gives the results of Adriot Manipulation tasks. Each curve represents the average total rewards with six random seeds. It is clear that TD3fG significantly outperforms other methods. DPGfD and SACfD fail to extract prior knowledge from the given trajectories. A possible explanation is that the samples are directly collected from humans via VR equipment, which may be unsuitable for robots and with low average rewards. Our approach uses demonstrations for exploration rather than directly training policies and shifts attention during training. TD3fG's exceptional performance validates its effectiveness for manipulator control.

\begin{figure}[h]
    \centering
    \subfloat[Ant]{\includegraphics[scale=0.11]{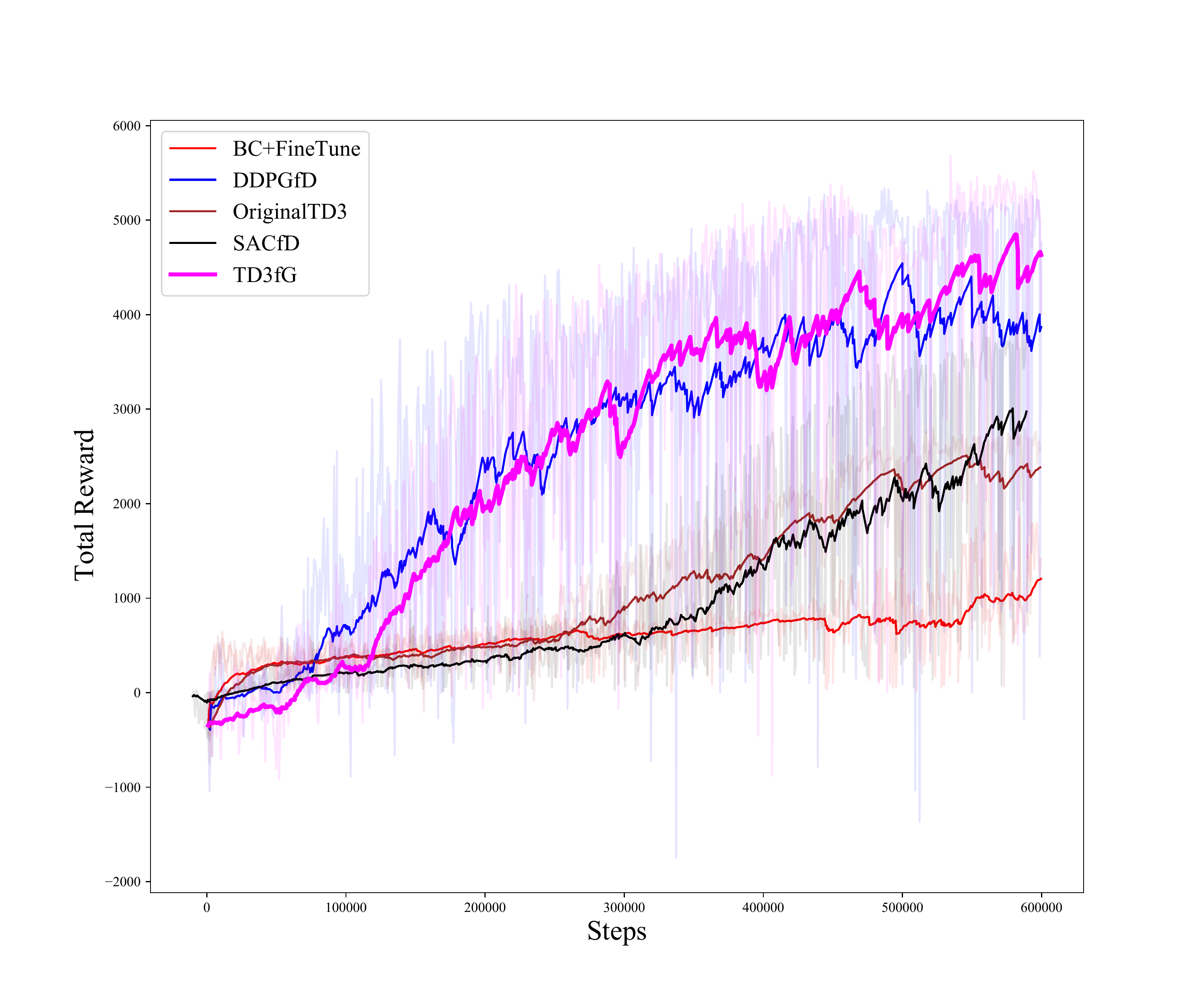}}
    \subfloat[HalfCheetah]{\includegraphics[scale=0.11]{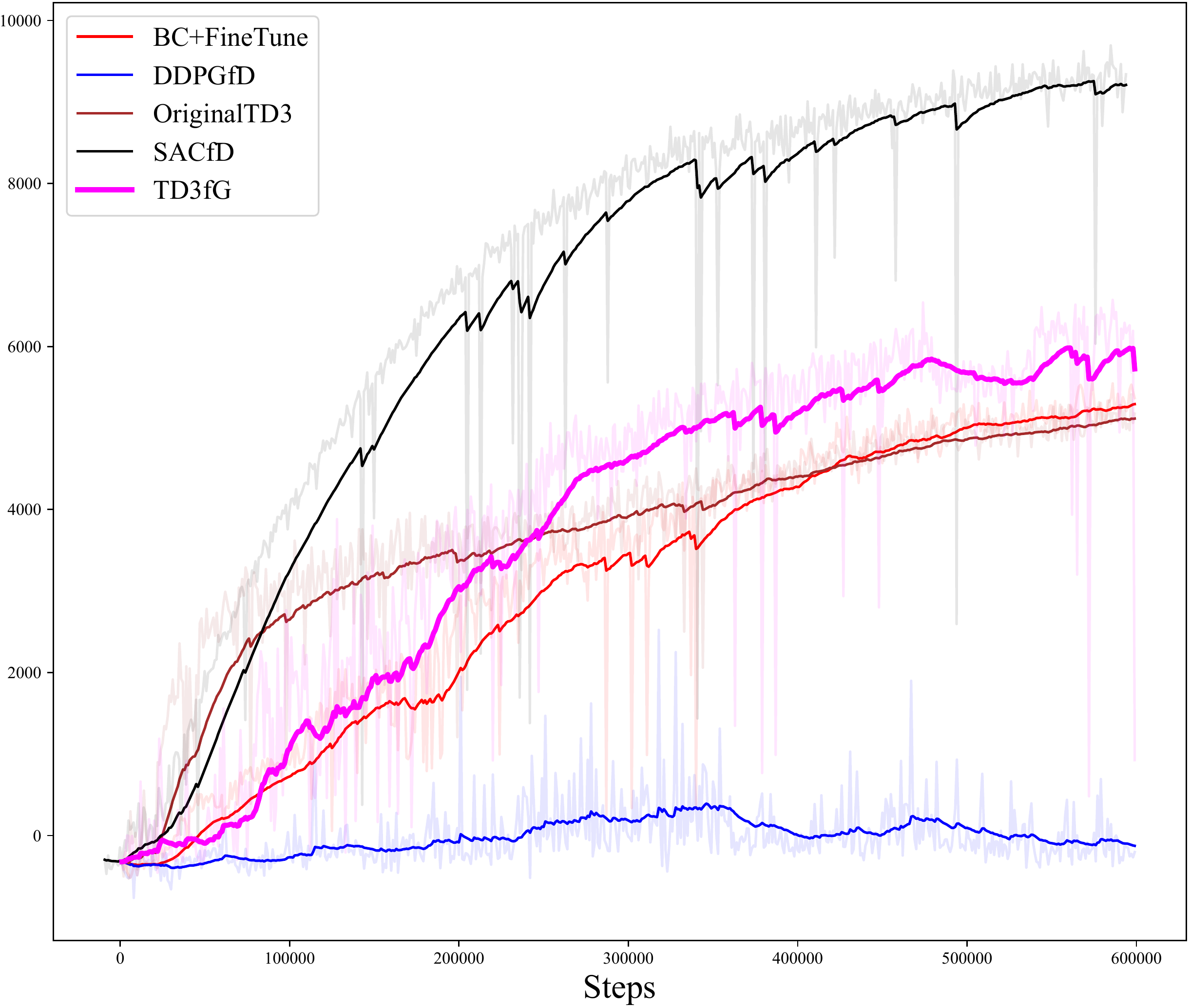}}
    \subfloat[Walker2d]{\includegraphics[scale=0.11]{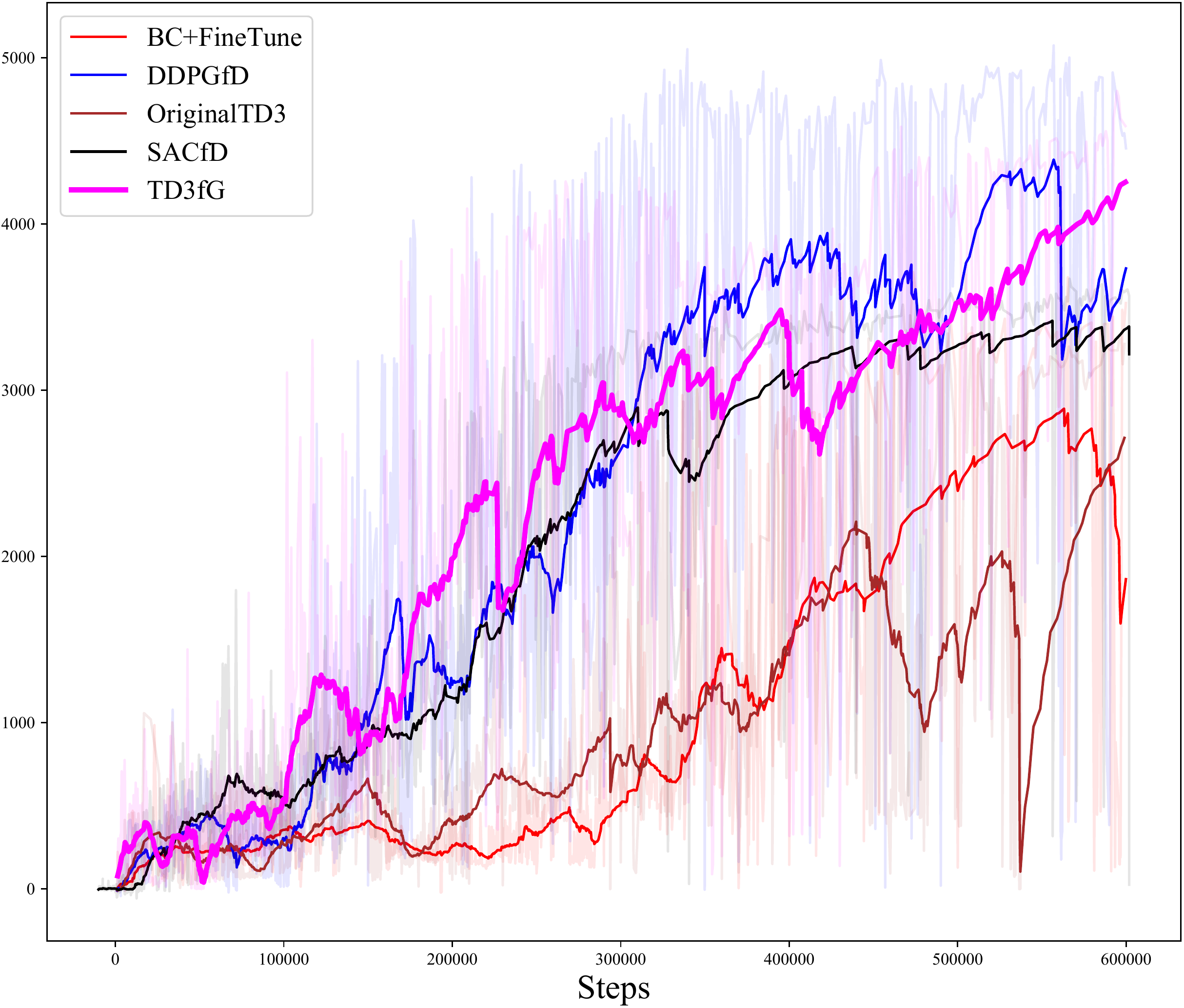}}
    \caption{Experiment results for MuJoCo tasks, where each curve presents the average total reward of 6 random seeds}
    \label{fig:Manipulator Compararison}
\end{figure}

To further analyze TD3fG, we extend it to MuJoCo tasks. The results of MuJoCo tasks are in Fig 4. It is hard for TD3 to finish MuJoCo without demonstrations. That may be due to the high dimension of action space and state space. Our method keeps a stable performance and exhibits 2x speeds up and 2x total rewards over the original TD3 in the Ant task. 

TD3fG performs best except in HalfCheetah is not as well as SAC. Previous studies suggest that SAC has a distinct advantage in this task\cite{haarnoja2018soft}\cite{haarnoja2018soft2}, but our approach worked better with higher training speed in the Ant and Walker2d. 

\begin{table*}[t]
\caption{comparative experimental results}
\centering
\label{table:results1}
\scalebox{1.0}{
\begin{tabular}{cccccc}  
\toprule   
    Tasks &hammer &door &Ant &HalfCheetah &Walker2d\\  
\midrule 
    Original TD3    &2628 &592 &2375.99    &5102.00    &2319.36\\  
    Behavior Cloning  &1861 &342 &3960.82    &5274.89    &1943.60\\    
    DDPGfD          &5283 &1248 &4194.04    &-115.85    &4213.81\\
    SACfD       &204.5 &762.3 &2873.46 &9241.35 &3013.68\\
    TD3fG         &9326.71 & 2345.04 &5023.12    &7902.64    &4065.45\\
  \bottomrule  
\end{tabular}
}
\end{table*}

The results reveal that DDPGfD and SACfD depend more on the training samples' quality, as the samples stored in the buffer will act on the entire training process. And their adaptability to different tasks varies considerably. However, our approach is well adapted to different environments and performs particularly well in the manipulator task. Meanwhile, it can leverage a reduced number of low-quality demonstrations and avoid adverse effects, which DDPGfD and SACfD cannot do.

Overall, by introducing a limited number of sub-optimal demonstrations, our approach outperforms other methods in manipulation and various robotics tasks. And it's more lenient on training samples.

\section{ABLATION EXPERIMENTS}

This section presents ablation experiments to evaluate the influence of different components.

Q-filter + BC loss: The BC loss will be imported if the Q value of reference action from generator $Q(s_t, \pi^{BC}(s_t)$ is bigger than $Q(s_t, \pi(s_t)$).

BC loss + Action noise: This method adds an exploration noise from the pre-trained generator, as in Section IV.

Demonstration Reply: We store 10000 demonstration transitions in the replay buffer. 

\begin{figure}[h]\small
    \centering
    \includegraphics[scale=0.11]{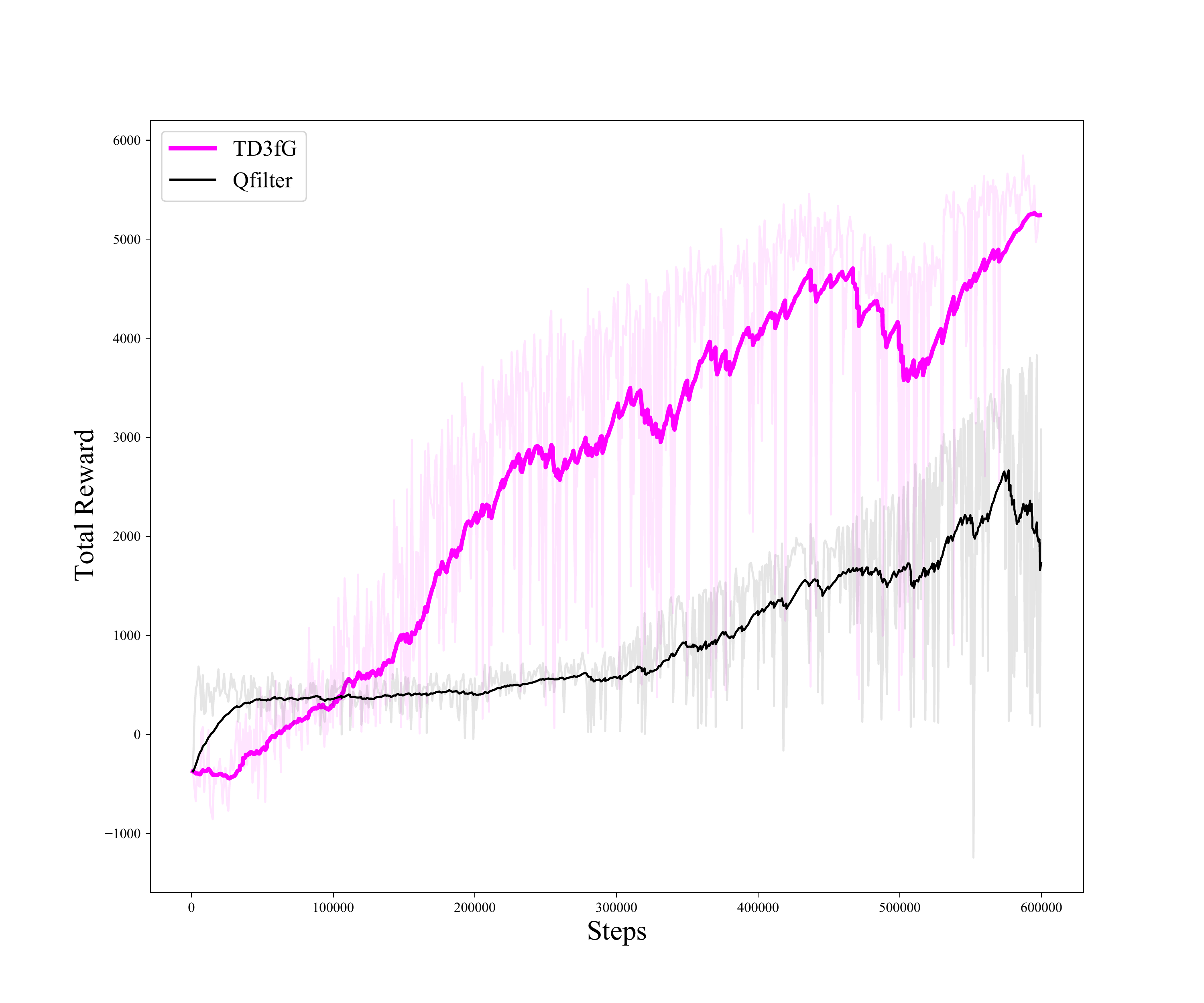}
    \includegraphics[scale=0.11]{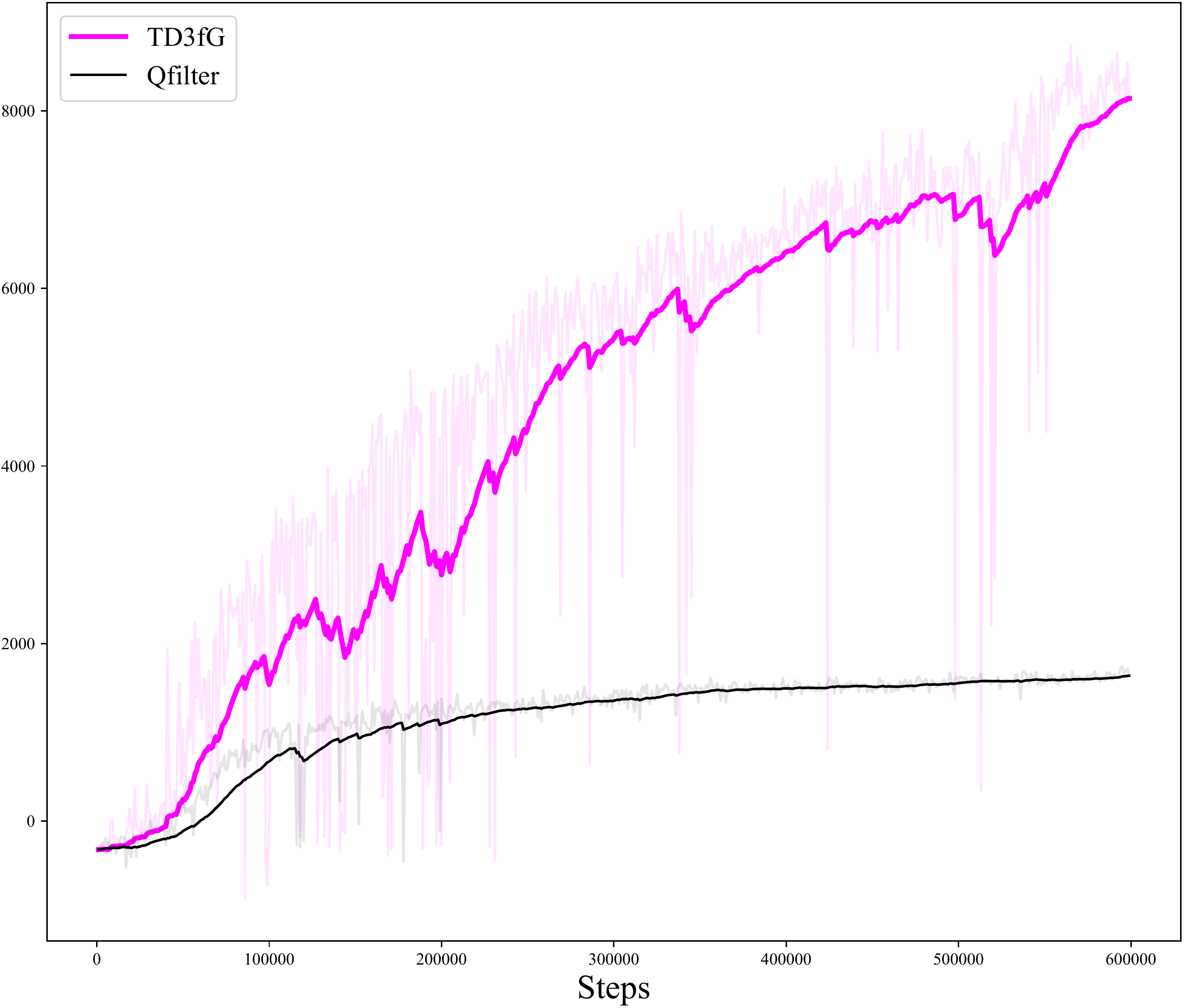}
    \includegraphics[scale=0.11]{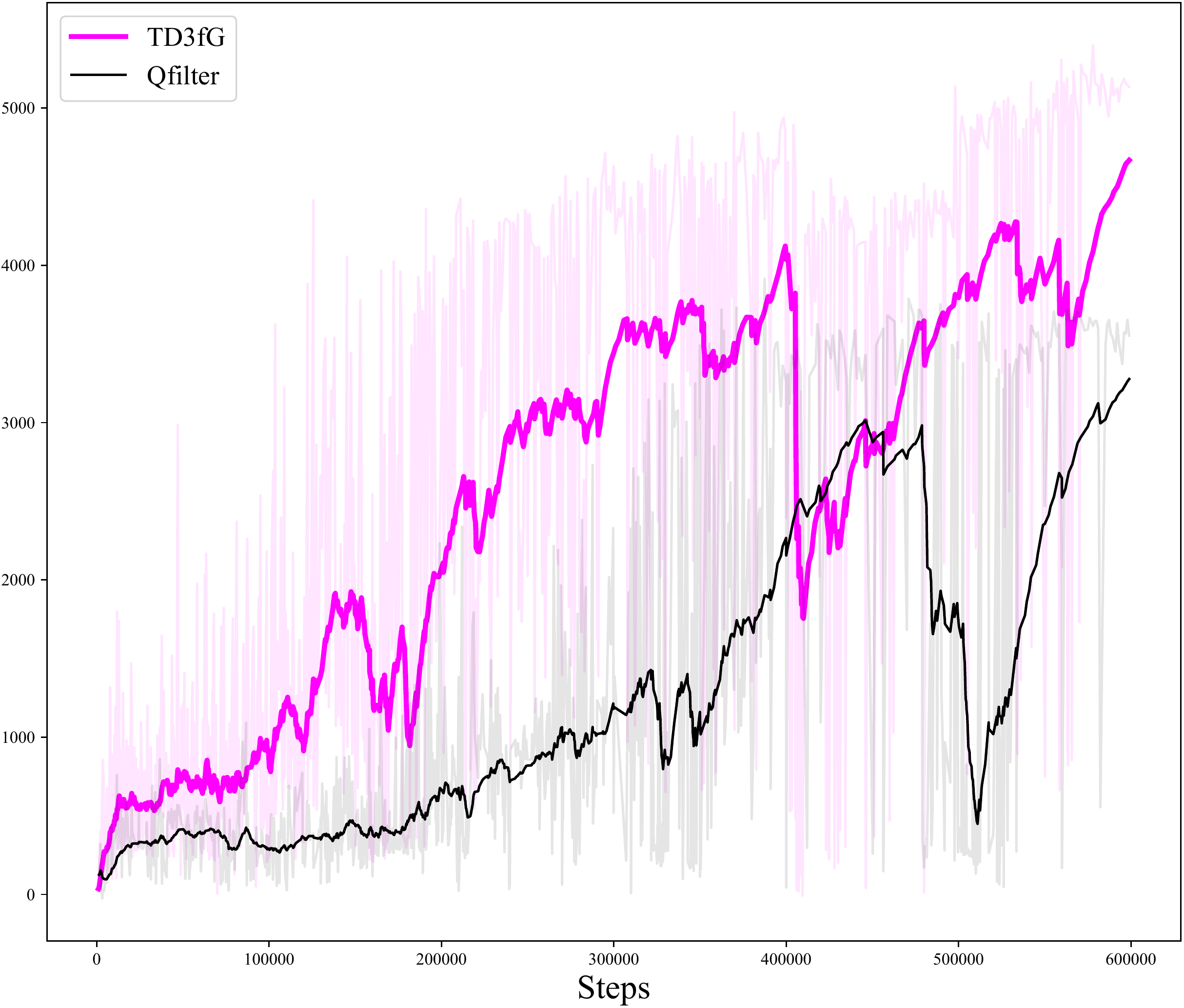}\\
    \includegraphics[scale=0.20]{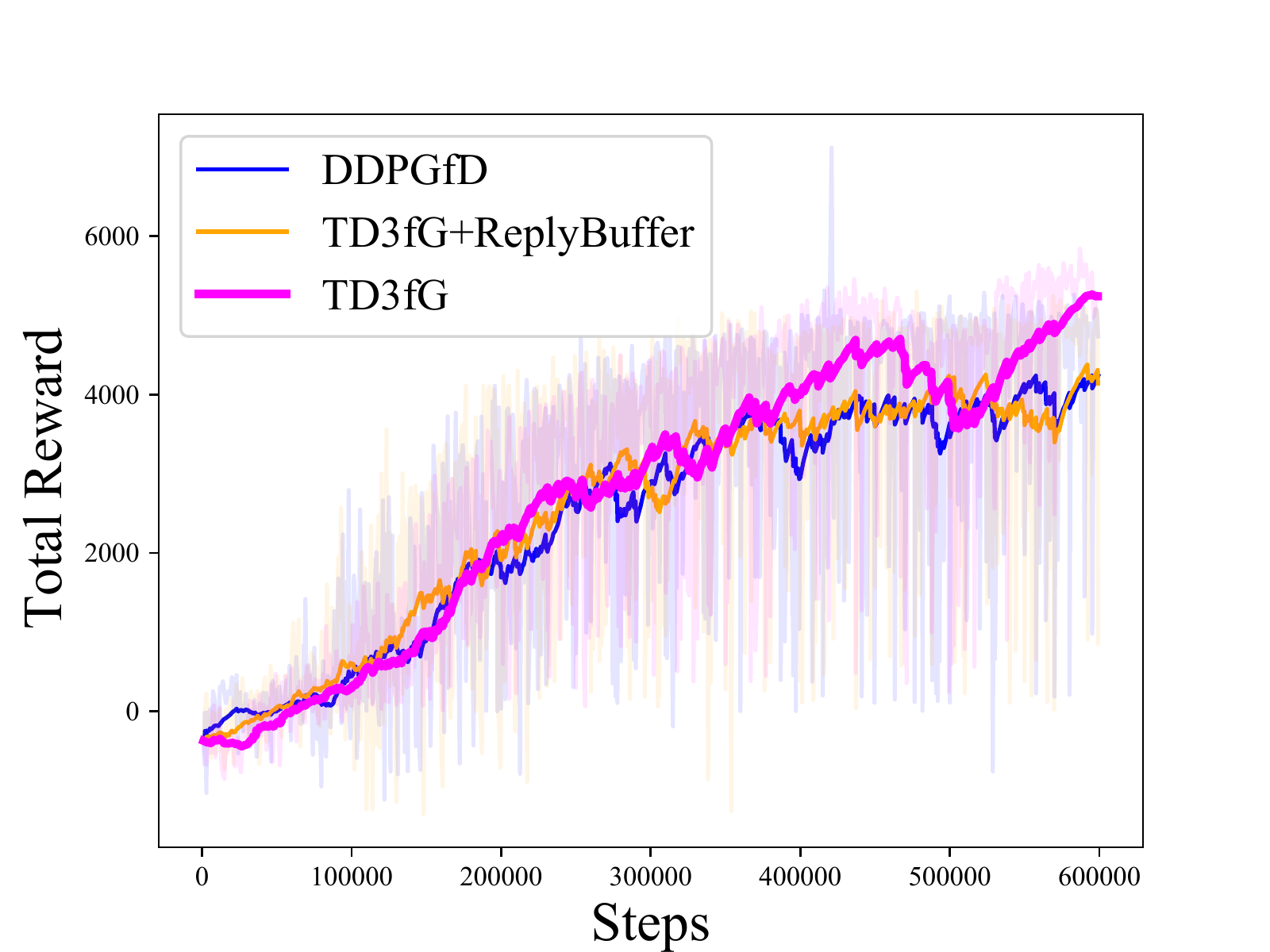}
    \includegraphics[scale=0.20]{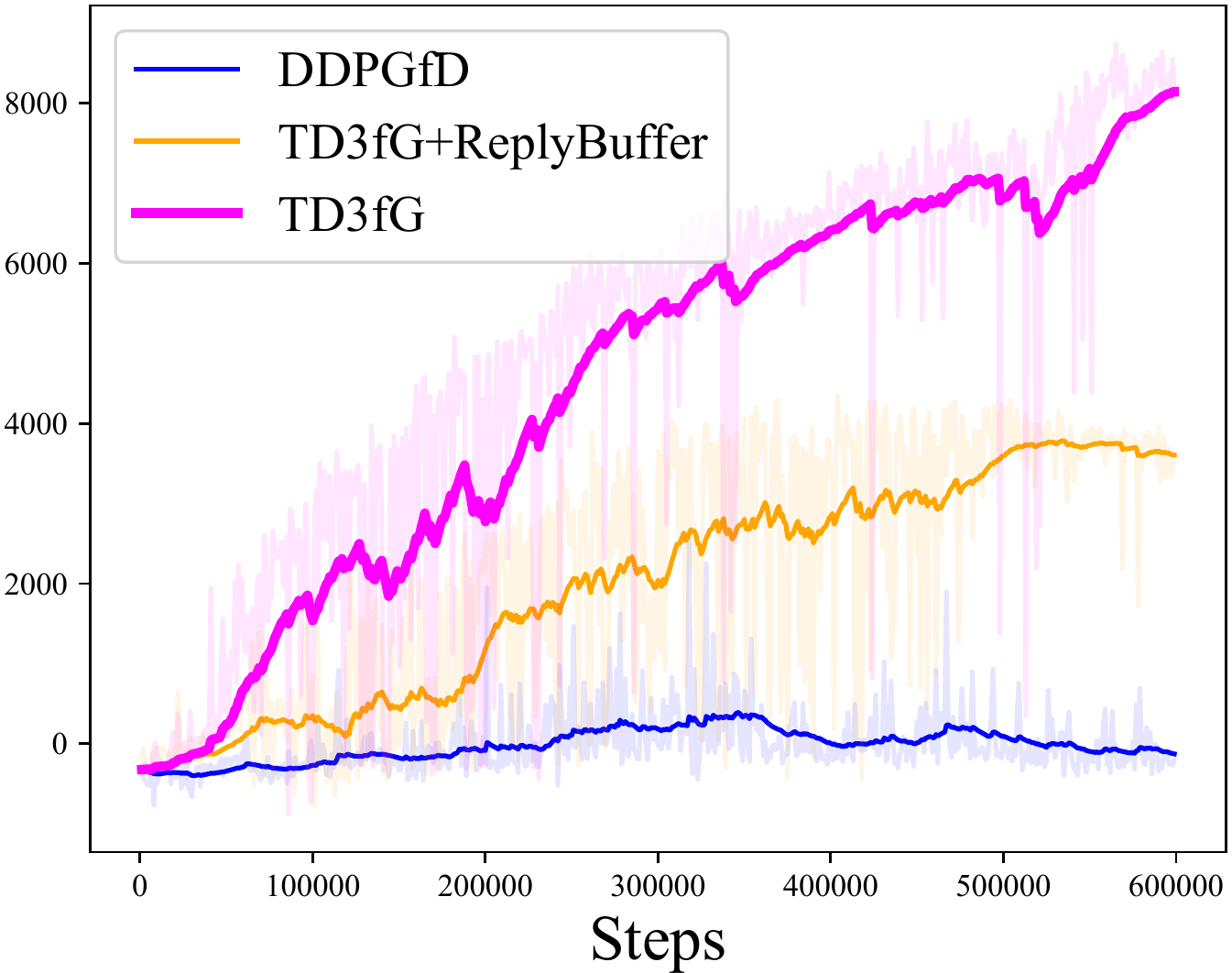}
    \includegraphics[scale=0.20]{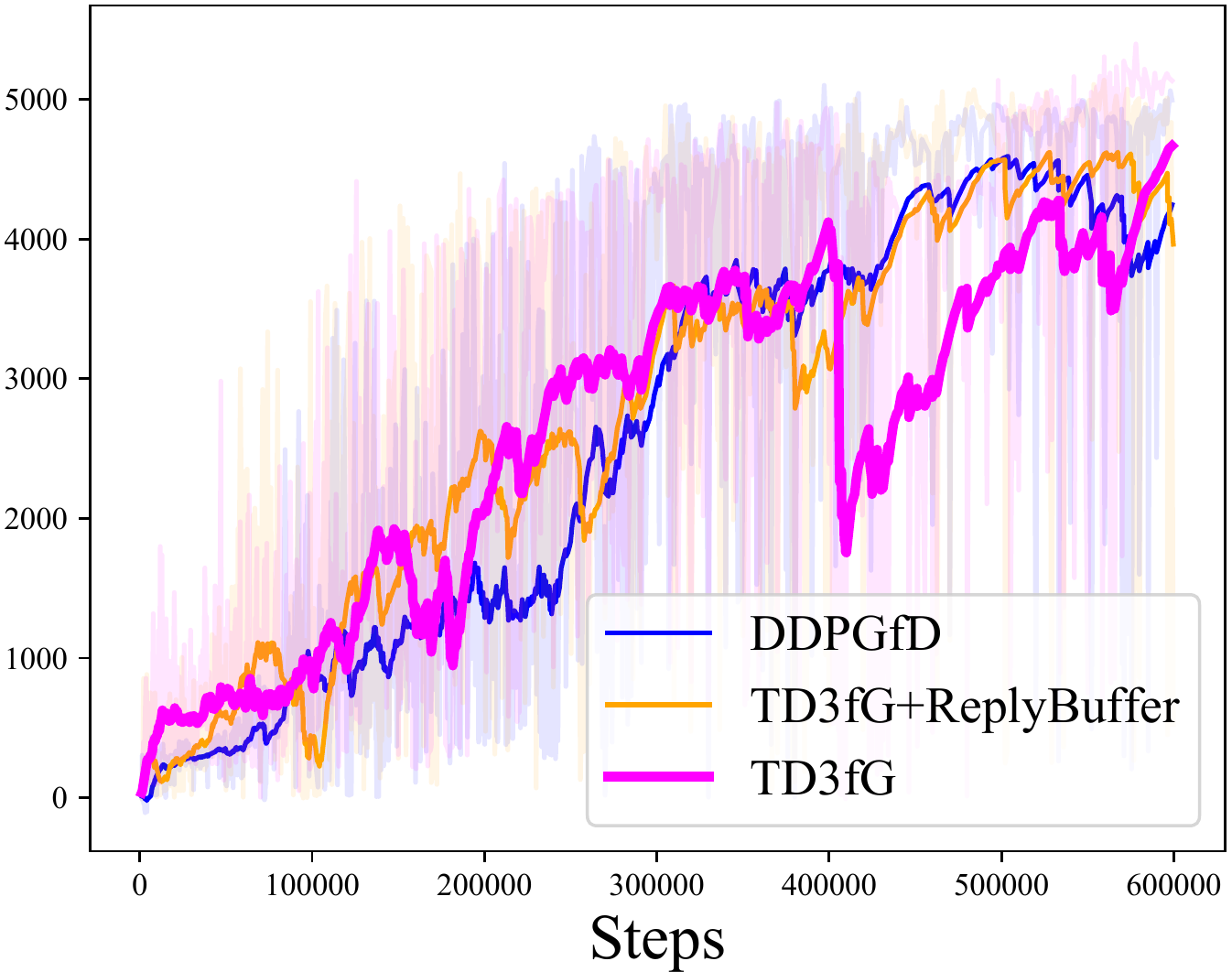}\\
    \subfloat[Ant]{\includegraphics[scale=0.20]{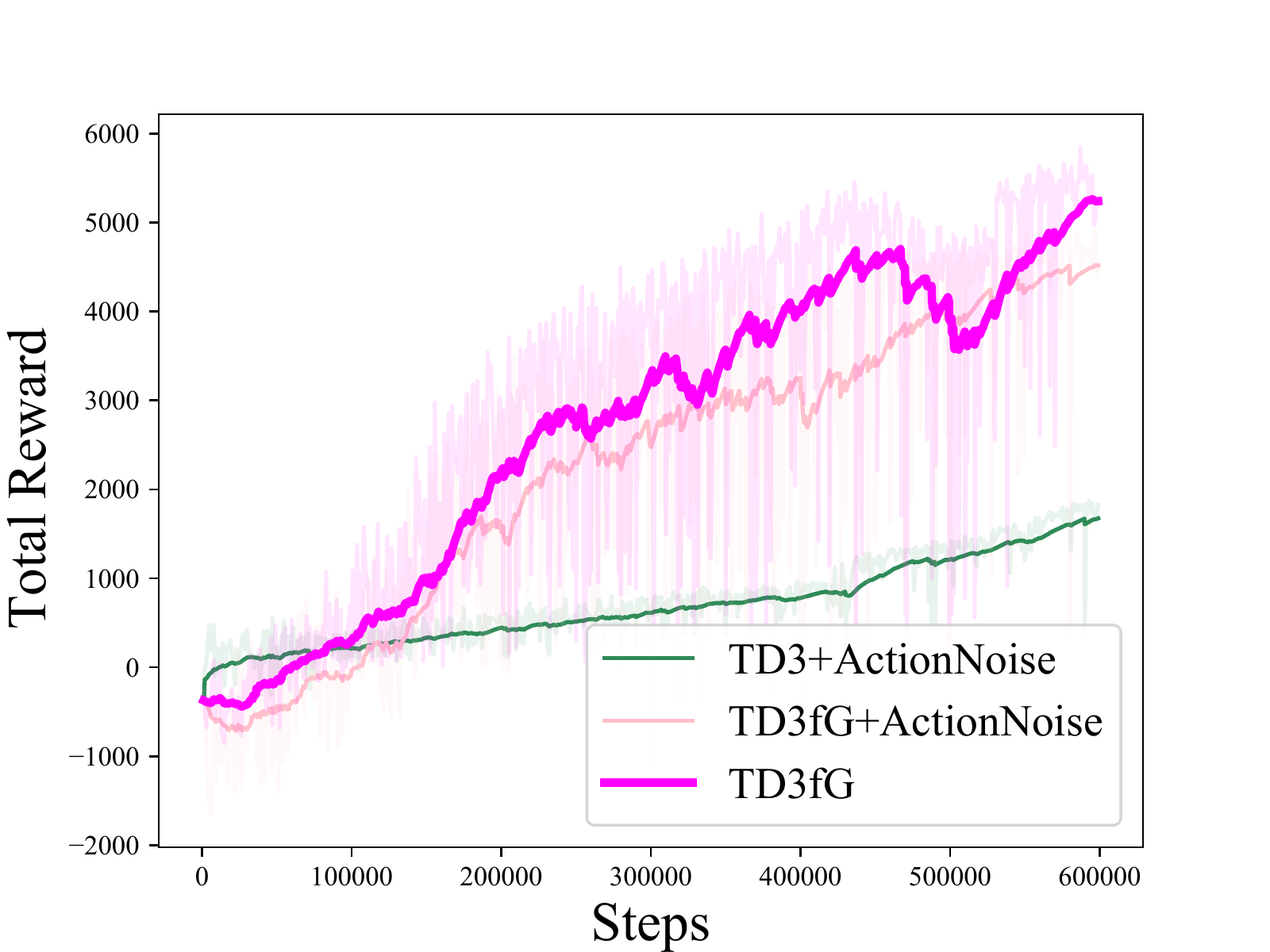}}
    \subfloat[HalfCheetah]{\includegraphics[scale=0.20]{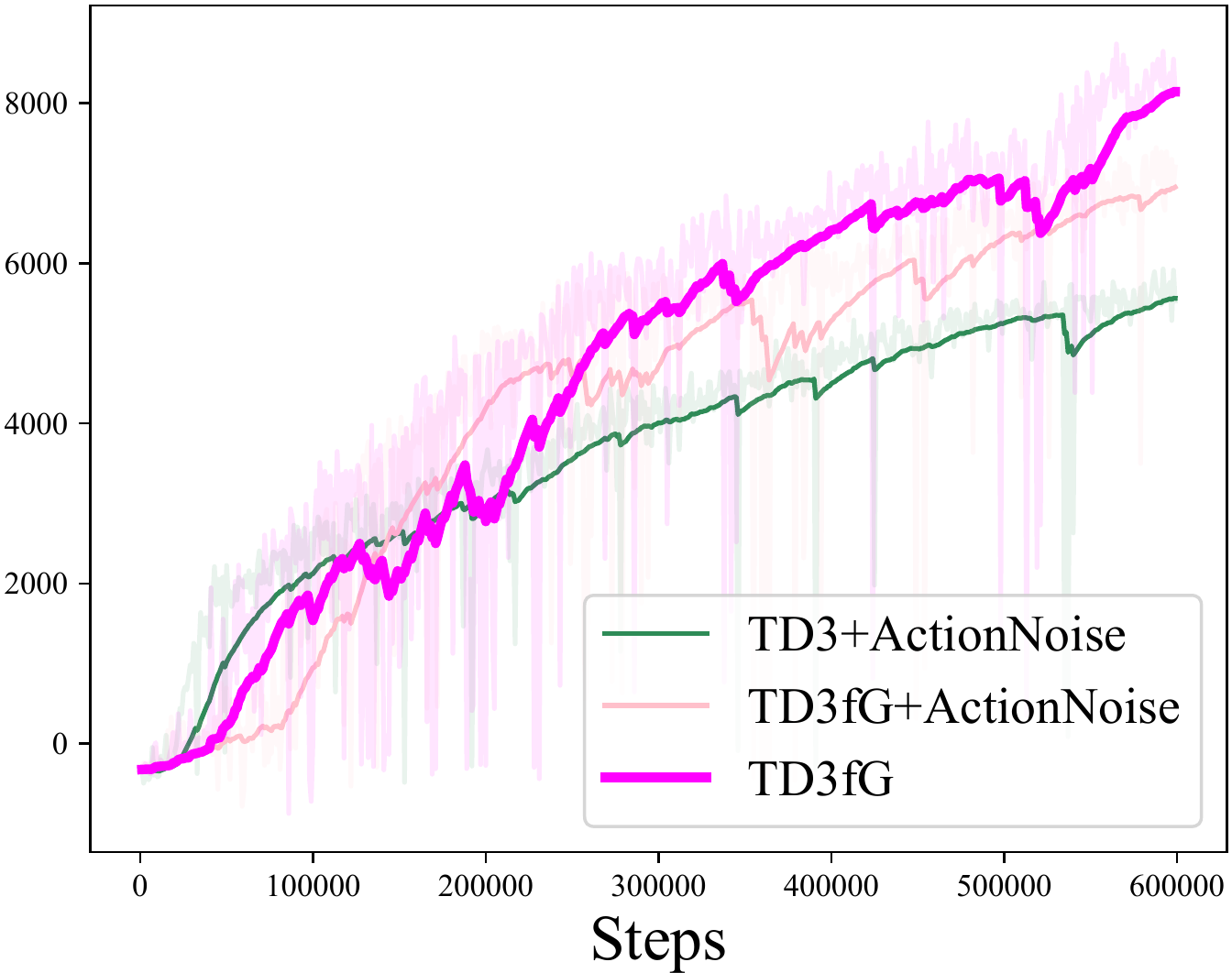}}
    \subfloat[Walker2d]{\includegraphics[scale=0.20]{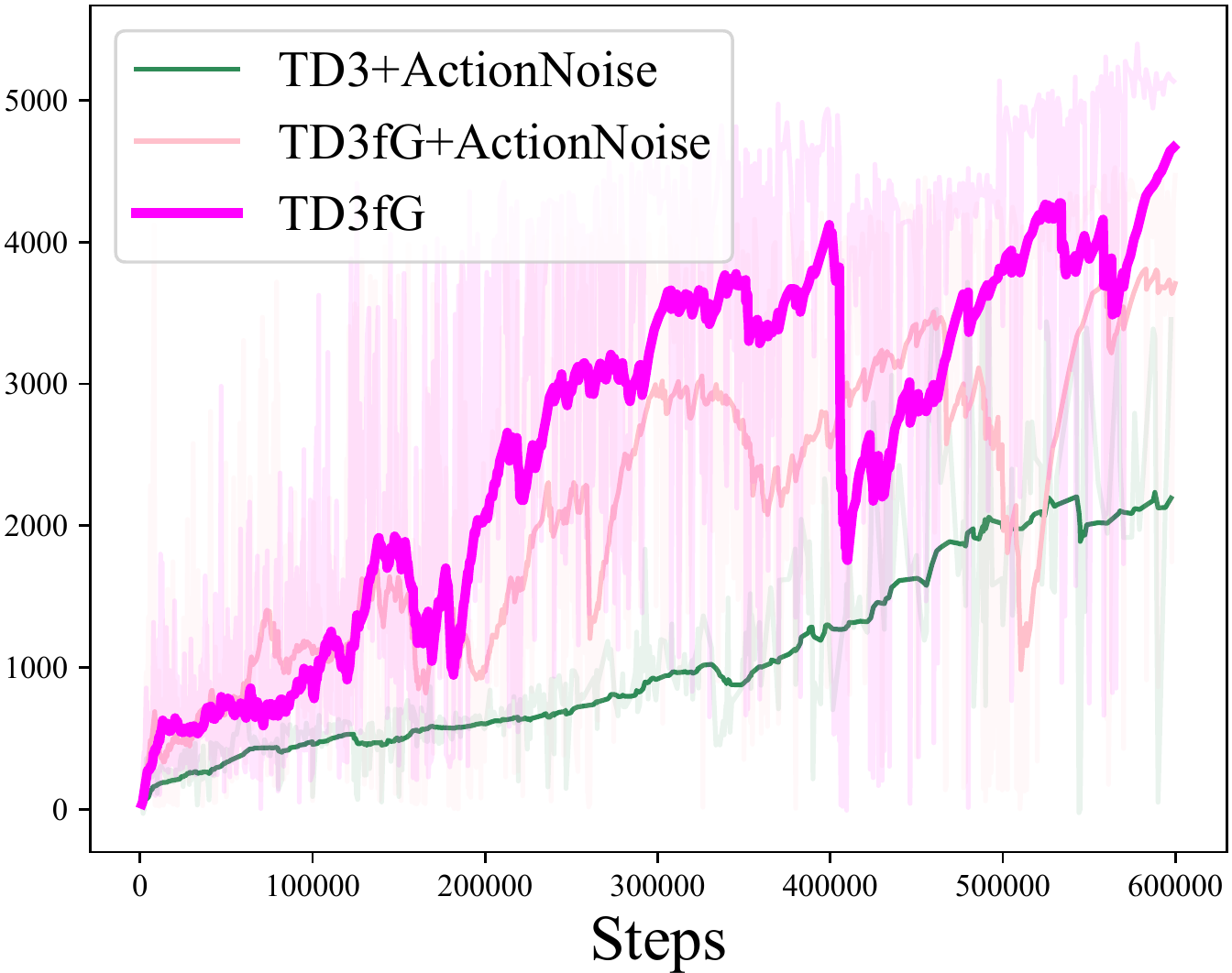}}
    \caption{Ablation experiment results. From top to bottom are Q-filter, Demonstration Reply, and Action Noise}
    \label{fig:my_label}
\end{figure}
\subsection{Q filter and Decreasing weight}

Another way to avoid over-concentrating on demonstrations is the Q-filter. The Q value of the reference action is obtained through the critic, then compared with the Q value of the actor's action. We only import the BC loss if the former is bigger.
\begin{equation}
\begin{aligned}
    & L^{G} = {1}_{Q(s_t, \pi^{G}(s_t)) > Q(s_t, \pi_{\phi_t}(s_t))} (a_t^{ref} - \pi_{\phi_t}(s_t))^2
\end{aligned}
\end{equation}

All results with the Q filter are worse, which can be attributed to two possible reasons. First, the Q filter lacks a smooth transition. Second, the critic's misevaluations in the early steps weaken the generator's guidance. The results substantiate that the proposed smooth transition is more helpful with suboptimal and limited demonstrations.
\begin{table}[h]
\caption{Comparison of TD3fG and Q filter}
\centering
\label{table_time}
\scalebox{1.0}{
\begin{tabular}{cccc}
\toprule   
    Tasks & Ant & HalfCheetah & Walker2d\\  
\midrule 
    TD3fG           &4704.44    &7601.85    &4065.45\\
    Q filter        &1809.60    &1631.40    &3155.78\\
  \bottomrule  
\end{tabular}
}
\end{table}

\subsection{Demonstration Reply}

Table IV exhibits the comparison results. The total reward lies clearly between DDPGfD and TD3fG. A major difference is a dependence on demonstrations, while TD3fG with demonstration replay is more sensitive to samples' quality than TD3fG. It coincides with our assumption about overreliance on demonstrations.
\begin{table}[h]
\setlength\abovedisplayskip{3pt}
\setlength\belowdisplayskip{3pt}
\centering
\caption{Comparison of Demonstration Reply}
\label{table:reply buffer}
\scalebox{1.0}{
\begin{tabular}{cccc} 
\toprule   
    Tasks & Ant & HalfCheetah & Walker2d\\  
\midrule 
    TD3fG               &4704.44    &7601.85    &4065.45\\
    TD3fG+Replay Buffer   &4110.68    &3615.32    &4047.55\\
  \bottomrule  
\end{tabular}
}
\end{table}

\subsection{Action noise}

In this part, we add an extra generated Action Noise (AN) and compare it with TD3fG, which only uses BC loss. Table V shows the rewards. Reference action alone boosts the results for all experiments but not as much as TD3fG with BC loss. 
\begin{table}[h]
\centering
\caption{Comparison of Action Noise}
\label{table_time}
\scalebox{1}{
\begin{tabular}{cccc}  
\toprule   
    Tasks & Ant & HalfCheetah & Walker2d\\  
\midrule 
    TD3fG               &4704.44    &7601.85    &4065.45\\
    AN   &1824.43    &5527.34    &2411.00\\
    TD3fG+AN &4494.95  &6913.96   &3662.40\\
\bottomrule  
\end{tabular}}
\end{table}

\section{DISCUSSION AND FUTURE WORK}
We propose a new RL with demonstrations algorithm that leverages samples in exploration noise and policy's lost function. TD3fG can extract prior knowledge and avoid poor samples' adverse effects with a smooth transition from learning from demonstrations to learning from experience. Our approach adapts well to limited samples and presents noticeable improvement in manipulator tasks. And it also works well in other robotics tasks. These findings help us to understand the role of unsuitable experts in RL with demonstrations.

A limitation of this study is that the generator's sensitivity to demonstrations remains further investigated. More data collection is required to determine exactly how the quality and quantity of samples affect the final performance and whether our idea can be extended to general robotics tasks.

We will further research how to leverage this kind of demonstration in offline settings, as there may not always be opportunities to interact fully with the environment. The decreasing confidence in demonstrations acts on all expert samples means the agent may ignore some high-quality samples that are worth learning. In future studies, we will train the agent to classify the demonstrations and use different learning methods to handle good trajectories and failure-biased trajectories separately.

\section{acknowledgement}
This research is supported by the Agency for Science, Technology, and Research (A*STAR) under its AME Programmatic Funding Scheme (Project \#A18A2b0046).

\bibliography{main}
\bibliographystyle{ieeetr}
\end{document}